\documentclass[conference]{IEEEtran}
\IEEEoverridecommandlockouts
\usepackage{cite}
\usepackage{amsmath,amssymb,amsfonts}
\usepackage{graphicx}
\usepackage{textcomp}
\usepackage{xcolor}
\def\BibTeX{{\rm B\kern-.05em{\sc i\kern-.025em b}\kern-.08em
    T\kern-.1667em\lower.7ex\hbox{E}\kern-.125emX}}

\usepackage{caption}
\usepackage{algorithm,algpseudocode}
\DeclareMathOperator*{\argmin}{\arg\min}
\DeclareMathOperator*{\argmax}{\arg\max}
\usepackage{bigstrut}
\usepackage{tabularx}
\usepackage{mathtools}
\usepackage{booktabs}
\usepackage{multirow}
\usepackage{subcaption}

\begin{document}

\title{Less is More: Understanding Word-level Textual Adversarial Attack via $\boldsymbol{n}$-gram Frequency Descend}


\author{\textbf{Ning Lu$^{1,2}$~\;~Shengcai Liu$^{3,5,*}$\thanks{$^*$Correspondence to  \texttt{liu\_shengcai@cfar.a-star.edu.sg}.}~\;~Zhirui Zhang$^4$~\;~Qi Wang$^5$~\;~Haifeng Liu$^6$~\;~Ke Tang$^1$}\\
  $^1$Guangdong Key Laboratory of Brain-Inspired Intelligent Computation, \\Department of Computer Science and Engineering, Southern University of Science and Technology~\;~\\
  $^2$Department of Computer Science and Engineering, Hong Kong University of Science and Technology~\;~\\
  $^3$Centre for Frontier AI Research, Agency for Science, Technology and Research~\;~$^4$Tencent AI Lab~\;~\\
  $^5$Department of Computer Science and Engineering, Southern University of Science and Technology~\;~\\
  $^6$OPPO Mobile Telecommunications Corp.
}



\maketitle

\begin{abstract}
Word-level textual adversarial attacks have demonstrated notable efficacy in misleading Natural Language Processing (NLP) models. Despite their success, the underlying reasons for their effectiveness and the fundamental characteristics of adversarial examples (AEs) remain obscure. 
This work aims to interpret word-level attacks by examining their $n$-gram frequency patterns.
Our comprehensive experiments reveal that in approximately 90\% of cases, word-level attacks lead to the generation of examples where the frequency of $n$-grams decreases, a tendency we term as the $n$-gram Frequency Descend ($n$-FD).
This finding suggests a straightforward strategy to enhance model robustness: training models using examples with $n$-FD. 
To examine the feasibility of this strategy, we employed the $n$-gram frequency information, as an alternative to conventional loss gradients,  to generate perturbed examples in adversarial training.
The experiment results indicate that the frequency-based approach performs comparably with the gradient-based approach in improving model robustness.
Our research offers a novel and more intuitive perspective for understanding word-level textual adversarial attacks and proposes a new direction to improve model robustness.
\end{abstract}

\begin{IEEEkeywords}
adversarial attack, natural language processing, AI safety
\end{IEEEkeywords}

\section{Introduction}
Deep Neural Networks (DNNs) have exhibited vulnerability to adversarial examples (AEs)~\cite{GoodfellowSS14, Papernot}, which are crafted by adding imperceptible perturbations to the original inputs.
In Natural Language Processing (NLP), numerous adversarial attacks have been proposed, which are typically categorized by the perturbation granularity: character-level~\cite{deepwordbug,textbugger}, word-level~\cite{GA, PWWS}, sentence-level~\cite{SCPNA, AddSent}, and mix-level modification~\cite{lei2022phrase}.
Among them, word-level attacks have attracted the most research interest, due to the superior performance on both attack success rate and AE quality~\cite{PruthiDL19,ta_benchmark}.
Thus, this work primarily explores these word-level attacks.
 
Simultaneously, the development of defenses against textual adversarial attacks has become a critical area of study. 
Notable defense strategies include adversarial training where the model gains robustness by training on the worst-case examples~\cite{Madry18, freelb, ascc},
adversarial data augmentation which trains models with AE-augmented training sets~\cite{pso}, AE detection~\cite{ScRNN,l2detect,spellchecker}, and certified robustness~\cite{ibp1,safer}.

\begin{figure}[t!]
    \centering
    \resizebox{1\linewidth}{!}{
    \begin{tabularx}{0.5\textwidth}{cX}
        \toprule
        \textbf{$n$-FD} & \multicolumn{1}{>{\centering\arraybackslash}X}{\textbf{Text}} \\
        \midrule
        Raw & it's \textbf{hard} to imagine that even very small children will \underline{be {\textbf{impressed}} by} this tired retread \\
        \midrule
        $1$-FD & it's $\overset{\textcolor{blue}{139 \rightarrow 16}}{\text{{\textbf{challenging}}}}$ to imagine that even very small children will be impressed by this tired retread \\
        \midrule
        $2$-FD & it's hard to imagine that even very small children will $\underset{\textcolor{red}{1\rightarrow 0, 4 \rightarrow 0}}{\underline{\text{be } \overset{\textcolor{blue}{6 \rightarrow 22}}{\textnormal{{\textbf{stunning}} }} \text{by}}}$ this tired retread \\
        
        \bottomrule
    \end{tabularx}
    }
    \caption{Illustrations of two AEs exhibiting 1-FD and 2-FD, respectively.
    The 1-gram (blue numbers) and 2-gram (red numbers) frequency changes are presented.
    In the second AE, the substitution of ``impressed'' with ``stunning'' raises the 1-gram frequency ($6 \rightarrow  22$). However, it concurrently reduces the 2-gram frequency ($1\rightarrow 0, 4 \rightarrow 0$).}
        
    \label{freq_eg}
\end{figure}

Despite the tremendous progress achieved, the fundamental mechanisms of word-level textual attacks, as well as the intrinsic properties of the AEs crafted by them, are not yet fully explored. 
Considering that textual attacks and defenses are generally oriented to security-sensitive domains such as spam filtering~\cite{bhowmick2018mail} and toxic comment detection~\cite{toxic_detect}, a clear understanding of textual attacks is important. It will elucidate the vulnerability of the DNN-based applications and contribute to enhancing their robustness.

This work seeks to understand word-level textual attacks from a novel perspective: $n$-gram frequency.
According to Zipf's law~\cite{zipflaw16}, the frequency of words (1-gram) in a linguistic corpus is generally inversely proportional to their ranks.
This means more common words appear exponentially more often than rarer ones, a pattern also holds true for $n$-grams ~\cite{cavnar1994n}.
While humans can easily navigate this frequency distribution disparity, DNNs struggle, which may lead to issues such as gender bias~\cite{machbias} and semantic blending~\cite{FRAGE}.
We hypothesize that the highly uneven distribution of $n$-grams may induce instability in models, particularly for sequences that occur less frequently, thus making them vulnerable to adversarial attacks.

To test this hypothesis, we thoroughly analyzed AEs generated by six different attack methods, targeting three DNN architectures across two dataset.
The results reveal a consistent pattern across all attacks: a strong tendency toward generating examples characterized by a descending $n$-gram frequency, i.e., AEs contain less commonly occurring $n$-gram sequences than original ones. 
Figure~\ref{freq_eg} showcases instances where AEs demonstrate decrease in $n$-gram frequency.
Moreover, this tendency is most pronounced when $n$ equals 2, broadening the earlier focus in this field that only considered the frequency of single words~\cite{freqdetect}.
Extra experiments also reveal that DNNs have difficulty processing $n$-FD examples.

These findings suggest a straightforward strategy to enhance model robustness: training on $n$-FD examples.
Unlike the common adversarial training approaches that use gradients to perturb examples to maximize loss, we suggest a new approach that perturbs examples to reduce their $n$-gram frequency.
We integrate this approach into the recent convex hull defense strategy~\cite{DNE} for adversarial training.
Surprisingly, our frequency-based approach performs comparably to the gradient-based approach in improving model robustness. 
In summary, our main contributions are:
\begin{itemize}

\item Our analysis reveals that word-level attacks exhibit a strong tendency toward generating $n$-FD examples.

\item Our experiments confirm that training models on $n$-FD examples can effectively improve model robustness, achieving defensive results comparable to the gradient-based approach.

\item We provide a novel, intuitive perspective for understanding word-level adversarial attacks through the lens of $n$-gram frequency. Additionally, we offer a new direction to enhance the robustness of NLP models by $n$-FD examples. 

\end{itemize}

\section{Understand Word-level Attacks from the $n$-FD Perspective}

In this section, we introduce the word-level textual attacks and the definition of $n$-gram frequency descend ($n$-FD).
Then we experimentally demonstrate that word-level adversarial attacks prefer generating $n$-FD examples. 

\subsection{Preliminaries}

\paragraph{Word-Level Textual Attacks}
As the most widely studied attacks in NLP~\cite{attacksurvey}, word-level textual attacks generate AEs by substituting words in the original texts.
Let $\boldsymbol{x} = [x_1, x_2, \cdots, x_L]$ denote a text with $L$ words. 
A word-level attack would first construct a candidate substitute set $S(x_i) = \{s_j^{(i)}\}_{j=0}^K$ with size $K$ for each word $x_i$.
Then it iteratively replaces a word in $\boldsymbol{x}$ with some substitute selected from the candidate set, until attack succeeds.

\paragraph{$n$-gram Frequency} By definition, $n$-gram is a contiguous sequence of $n$ words in the given texts.
The $i$-th $n$-gram of text $\boldsymbol{x}$ is defined as: 
\begin{equation}
    \label{eq: ngram}
    \begin{aligned}
    g_{i}^n \coloneqq [x_i, x_{i+1}, \cdots, x_{i + n - 1}], i \in [1, L - n + 1].
    \end{aligned}
\end{equation}
We define the number of occurrences of an $n$-gram in the training set as its $n$-gram frequency, denoted as $\phi(g_{i}^n)$.
Then the $n$-gram frequency of text $\boldsymbol{x}$, denoted as $\Phi_n(\boldsymbol{x})$, is the average of the $n$-gram frequencies of its all $n$-grams:
\begin{equation}
\label{eq:freq_x}
    \Phi_n(\boldsymbol{x}) \coloneqq \frac{1}{L - n + 1} \sum_{i=1}^{L - n + 1} \phi(g_{i}^n).
\end{equation}

\paragraph{$n$-gram Frequency Descend ($n$-FD)}
Given $\boldsymbol{x}$, supposing an attack generates an example  $\boldsymbol{x}'$ by substituting some words in $\boldsymbol{x}$, then $\boldsymbol{x}'$ is a $n$-FD example if it has lower $n$-gram frequency than $\boldsymbol{x}$, i.e., $\Phi_n(\boldsymbol{x'}) < \Phi_n(\boldsymbol{x})$.

Similarly, $\boldsymbol{x}'$ is a $\boldsymbol{n}$-gram frequency ascend ($\boldsymbol{n}$-FA) example and a $\boldsymbol{n}$-gram frequency constant ($\boldsymbol{n}$-FC) example if $\Phi_n(\boldsymbol{x'}) > \Phi_n(\boldsymbol{x})$ and $\Phi_n(\boldsymbol{x'}) = \Phi_n(\boldsymbol{x})$, respectively.

\begin{table}[tbp]
    \centering
     \caption{Summary of the attacks for AE generation, including model access, substitution methods, and search strategies. W/B represents white/black-box attack. WSG represents word-saliency-based greedy search.}
    \resizebox{0.9\linewidth}{!}{
      \begin{tabular}{cccc}
      \toprule
      Attack & Access   & Substitution & Search \\
      \midrule
      GA~\cite{GA}    & B     & Counter-fitted & Genetic  \\
      PWWS~\cite{PWWS}  & B     & WordNet & WSG \\
      TF~\cite{textfooler}    & B     & Counter-fitted & WSG \\
      PSO~\cite{pso}   & B     & HowNet & Particle Swarm \\
      LS~\cite{localsearch}   & B     & HowNet & Local Search \\
      HF~\cite{hotflip}    & W     & Counter-fitted  & Gradient \\
      \bottomrule
      \end{tabular}%
    }
    \label{tab:attacks}%
\end{table}%

\paragraph{$n$-FD Substitution}
If a word substitution decreases the $n$-gram frequency of the text, then it is dubbed $n$-FD substitution.
Formally, given text $\boldsymbol{x}$, let $\boldsymbol{x}_{x_i \rightarrow s_j^{(i)}}$ denote the text generated by substituting $x_i$ in $\boldsymbol{x}$ with $s_j^{(i)}$.
Then the $n$-gram frequency change of the text, denoted as $\Delta \Phi_{n}(s_j^{(i)};\boldsymbol{x})$, is:
\begin{equation}
\label{eq:freq_delta}
    \Delta \Phi_{n}(s_j^{(i)};\boldsymbol{x}) \coloneqq \Phi_n(\boldsymbol{x}_{x_i \rightarrow s_j^{(i)}}) - \Phi_n(\boldsymbol{x}).
\end{equation}
If $\Delta \Phi_{n}(s_j^{(i)};\boldsymbol{x}) < 0$, then substitution $x_i \rightarrow s_j^{(i)}$ is a $n$-FD substitution.
Similarly, if $\Delta \Phi_{n}(s_j^{(i)};\boldsymbol{x}) > 0$ and $\Delta \Phi_{n}(s_j^{(i)};\boldsymbol{x}) = 0$, then it is a $n$-FA substitution and a $n$-FC substitution, respectively. 
For example, in Figure~\ref{freq_eg}, the replacement of \emph{``hard''} $\rightarrow$ \emph{``challenging''} is a 1-FD substitution, while \emph{``impressed''} $\rightarrow$ \emph{``stunning''} is a  2-FD substitution.



\subsection{Adversarial Example Generation}

\paragraph{Attacks} 
We selected six existing word-level attacks, including five black-box attacks: GA~\cite{GA}, PWWS~\cite{PWWS}, TextFooler (TF)~\cite{textfooler}, PSO~\cite{pso}, LocalSearch (LS)~\cite{localsearch}; and one white-box attack: HotFlip (HF)~\cite{hotflip}.
These attacks are representative in the sense that and have achieved effective performance against various models.
They employs different substitute candidate construction methods and search strategies, as summarized in Table \ref{tab:attacks}. 
To construct the substitute sets, PSO and LS use the language database of HowNet~\cite{hownet}, PWWS uses WordNet~\cite{wordnet}, while GA, TF, and HF rely on Counter-fitted~\cite{CF} embeddings.



\paragraph{Dataset}
We ran these attacks on 1000 test examples randomly selected from two public classification datasets: IMDb reviews dataset (IMDB)~\cite{IMDB} for sentiment analysis and AG-News corpus (AGNews)~\cite{AGnews} for topic classification. 

\paragraph{Victim Models}
The attacking experiments covered three different DNN architectures: convolutional neural network (CNN)~\cite{wordcnn}, long short-term memory (LSTM)~\cite{lstm} and pre-trained BERT~\cite{bert}. 


\begin{figure*}[tbp]
    \centering
    \includegraphics[width=\textwidth]{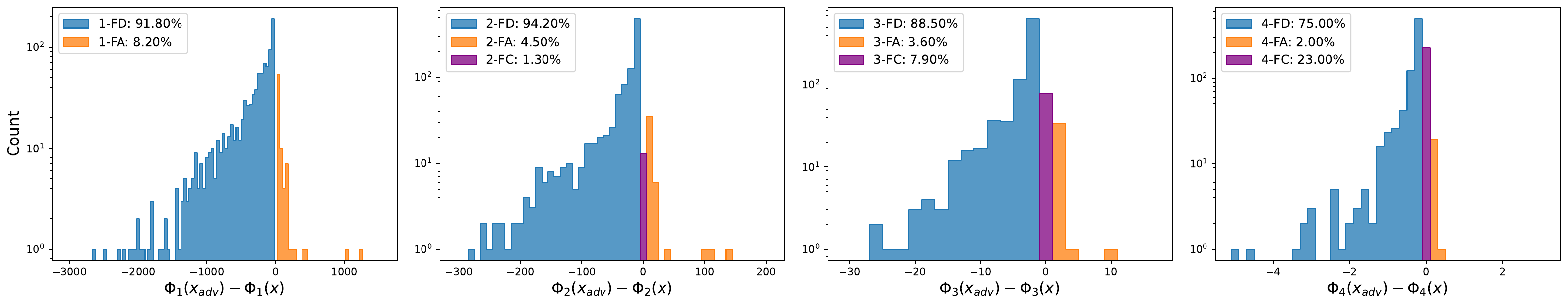}
    \caption{Distributions of the $n$-gram frequency changes induced by PWWS attack when attacking BERT on the IMDB dataset.
    The blue, orange, and purple bars represent the $n$-FD, $n$-FA, and $n$-FC examples, respectively.
    The exact percentage values are shown in the legend. 
    From left to right, the value of $n$ varies from 1 to 4.
    }
    \label{fig:sub dist} 
\end{figure*}

\begin{table}[tbp]
    \centering
    \caption{Percentages of the $n$-FD, $n$-FA, and $n$-FC examples in the AEs generated by all the six attacks when attacking three models on two datasets.
    }
    \resizebox{0.68\linewidth}{!}{
      \begin{tabular}{cccc}
\toprule
$n$   & $n$-FD (\%) & $n$-FC (\%) & $n$-FA (\%) \\
\midrule
1     & 91.27 & 0.75  & 7.98 \\
2     & 93.51 & 2.58  & 3.92 \\
3     & 87.29 & 10.55 & 2.17 \\
4     & 72.56 & 26.24 & 1.21 \\
\bottomrule
\end{tabular}%
}
    \label{tab:nfd_rate}%
\end{table}%

\subsection{Results and Analysis}
\label{sec:attack result}

\textbf{Adversarial attacks have $n$-FD tendency.} Table~\ref{tab:nfd_rate} summarizes the average percentages of $n$-FD, $n$-FA, and $n$-FC examples (with $n=1,2,3,4$) of the AEs generated by the six attacks, calculated across three models and two dataset.
One can observe that when $n$ is $1, 2, 3$, around 90\% of the AEs show $n$-FD characteristic.
Figure~\ref{fig:sub dist} further shows the detailed distributions of the $n$-gram frequency changes induced by the PWWS attack.
It can be observed that most changes show the decreases in terms of $n$-gram frequency.
Overall, all these attacks exhibit a strong tendency toward generating $n$-FD examples.

\begin{figure*}[t!]
    \centering
    \includegraphics[width=\textwidth]{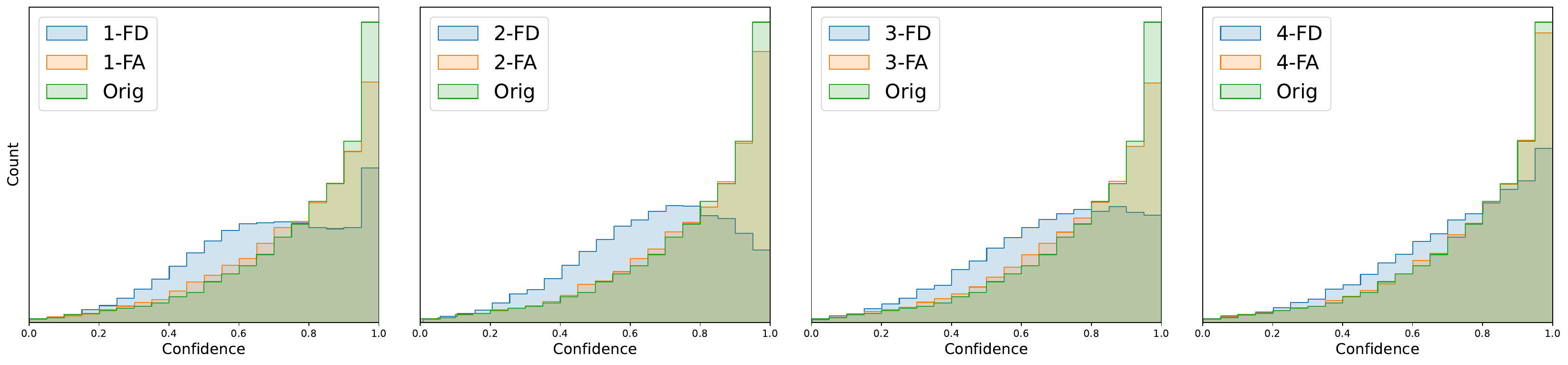}
    \caption{The confidence distribution for a CNN classifier on the clean examples (Orig), $n$-FD examples ($n$-FD) and $n$-FA examples ($n$-FA) from IMDB dataset. 
    Confidence refers to the softmax probability of the true class.
    $n$ is 1 to 4 from left to right images.
    Models perform similarly on clean examples and $n$-FA examples, but worse on $n$-FD examples. 
    }
    \label{fig:conf_dist} 
\end{figure*}

\textbf{$n$-FD tendency is most
pronounced when $n$ equals 2.}
Based on Table~\ref{tab:nfd_rate}, $2$-FD examples achieve better coverage compared with other cases. 
When $n = 1$, the percentage of $n$-FA is high, indicating a significant portion of the AEs containing more-frequent words.
For $n = 3, 4$, the percentage of $n$-FC is large, which means they are not good indicator to interpret AEs.
Further experiments show that, on average, 97\% of $n$-FC cases are out-of-vocabulary (OOV) replacement, where both original and new $n$-grams never appear in the training set. 

\textbf{Models exhibit reduced performance on $n$-FD examples.} 
Previous analysis implies that NLP models struggle more with  $n$-FD examples.
To test this hypothesis, we conducted experiments using the IMDB test set.
For each test example, we generated one $n$-FD and one $n$-FA example through random word substitutions. 
Then, we evaluated the standardly trained models on the three sets: the original test set, $n$-FD example set, and $n$-FA example set.
Figure~\ref{fig:conf_dist} shows that the model's predictions are less accurate on $n$-FD examples than on $n$-FA examples.
This outcome is expected, as DNNs typically do not learn effectively from small sample sizes without specific training techniques. 
Besides, the poor adaption to smaller samples has a minor effect on evaluation metrics, which are designed to assess performance across a broad range of data.

\section{Training on $n$-FD Examples Improves Robustness}
\label{sec:nFD training}

The findings from the previous section indicate that AEs  exhibit an $n$-FD tendency, on which models perform poorly. 
Building upon this insight, a intuitive approach is to train models on $n$-FD examples, similar to adversarial training.
To evaluate the feasibility of this approach, we  developed an adversarial training framework that relies on $n$-gram frequency.
In this approach, it's the frequency, not the gradient, that directs the generation of AEs.
This section will detail the approach.


\subsection{$n$-FD Adversarial Training}
In the conventional adversarial training paradigm, the training object is modeled as a min-max problem, where the inner goal is to find an example $x_{adv}$ that maximizes the prediction loss, formulated as:
\begin{equation}
 \boldsymbol{x}_{adv} = \argmax_{\boldsymbol{x}' \in \mathcal{P}(\boldsymbol{x})} \mathcal{L}(F(\boldsymbol{x}'), y),
\end{equation}
where $\boldsymbol{x}_{adv}$ denotes the loss maximizing example, and $\mathcal{P}(\boldsymbol{x})$ is the perturbation set consisting of all texts that can be generated by doing the substitution operation on $\boldsymbol{x}$.
$\mathcal{L}$ denotes the loss function of trained model $F$.
In practice, training algorithm literately update $\boldsymbol{x}$ to approximately approach $\boldsymbol{x}_{adv}$ with the help of gradient.

To access the effectiveness of $n$-FD examples, we modify the gradient-based adversarial training paradigm to $n$-FD adversarial training, where the inner objective is to find the $n$-FD example, formulated as:
\begin{equation}
    \boldsymbol{x}_{n-\text{FD}} = \argmin_{\boldsymbol{x}' \in \mathcal{P}(\boldsymbol{x})} \Phi_n(\boldsymbol{x}').
    \label{eq:nFD training}
\end{equation}

\begin{algorithm}[t]
\caption{$n$-FD convex hull training}
\label{alg:nfd CV}
\begin{algorithmic}[1]
\Require Dataset $D$, model $f$ with parameter $\theta$, adversarial steps $T_{adv}$
\State Initialize $\theta$ 
\State Initialize $n$-gram frequency table $T_{\Phi_n}$.
\For{epoch = $1 \cdots N_{epoch}$}
    \For{$\boldsymbol{x}, y \in D$}
        \State Randomly initialize $\boldsymbol{w}_{0}$
        \State $\boldsymbol{g} \leftarrow 0$
        
        \For{$t = 0$ to $T_{adv}$}
            \State Compute $\boldsymbol{\widetilde{x}}_t$ by Eq. (\ref{convexhull}) using $\boldsymbol{w}_{t}$
            \State $\boldsymbol{g} \leftarrow \boldsymbol{g} +
            \nabla_{\theta} L(f(\widetilde{\boldsymbol{x}}_t), y) $
            
            \State Update n-gram frequency table $T_{\Phi_n}$ by $\widetilde{\boldsymbol{x}}_t$
            \State Compute $\boldsymbol{w}_{t+1}$ by Eq.~(\ref{eq:nfd w}) using $T_{\Phi_n}$.
            
        \EndFor
        \State Update $\theta$ by $g$
    \EndFor
\EndFor

\end{algorithmic}
\end{algorithm}

\subsection{Applying $n$-FD Adversarial Training to Convex Hull} 

\paragraph{Convex Hull Framework}
We apply our method to a recently proposed convex hull paradigm~\cite{DNE, ascc}. 
The generated AE during training is a sequence of virtual vectors $\widetilde{x_i}$, which is a convex combination of synonyms $S(x_i)$, formulated as: 
\begin{equation}
\label{convexhull}
    \widetilde{x_i} = \sum_{j=0}^{K} w^{(i)}_j s^{(i)}_j,
\end{equation}
where the $w^{(i)}_j$ is the corresponding coefficient for substitution $s_j^{(i)}$, which meet the convex hull constraints $\{ \sum_j w^{(i)}_j = 1, w^{(i)}_j > 0\}$.
Previous works used gradient-based methods to update $w$ and build loss-maximizing AEs during training, formulated as:
\begin{equation}
    \Delta w_j^{(i)} = \alpha \lVert \nabla_{w_j^{(i)}} \mathcal{L} \rVert ,
\end{equation} 
where $\alpha$ is adversarial step size and $\Vert . \Vert$ represents $l$-$2$ normalize operation.

\begin{table*}[htbp]
  \centering
  \caption{Classification accuracy (\%) of ADV-G and ADV-F on clean examples (CLN) and different attacks across models and dataset. ``AVG ROB'' represents the average defensive performance against various attacks, with the numbers in brackets indicating the performance difference between ADV-F and ADV-G.}
    \begin{tabular}{ccccccccccccccl}
    \toprule
    \multirow{2}[4]{*}{Dataset} & \multirow{2}[4]{*}{Defense} & \multicolumn{4}{c}{CNN}       & \multicolumn{4}{c}{LSTM}      & \multicolumn{4}{c}{BERT}      & \multirow{2}[4]{*}{AVG ROB} \\
\cmidrule(lr){3-6} \cmidrule(lr){7-10}  \cmidrule(lr){11-14}          &       & CLN   & PWWS  & TF    & LS    & CLN   & PWWS  & TF    & LS    & CLN   & PWWS  & TF    & LS    &  \\
    \midrule
    \multirow{4}[2]{*}{IMDB} & Standard & 88.7  & 2.3   & 5.8   & 2.0   & 87.4  & 2.4   & 8.2   & 0.9   & 94.1  & 34.7  & 36.2  & 3.5   & 10.7  \\
          & ADV-G & 83.8  & 70.0  & 71.6  & 69.6  & 83.5  & 72.4  & 73.5  & 70.1  & 92.9  & 63.4  & 65.0  & 58.5  & 68.2  \\
          & ADV-F1 & 87.3  & 69.8  & 71.2  & 68.8  & 87.0  & 68.0  & 69.2  & 66.4  & 93.2  & 66.1  & 67.3  & 63.3  & 67.8 {\tiny ($\mathrel{-}$0.4)}  \\
          & ADV-F2 & 87.6  & 70.2  & 72.0  & 70.5  & 86.5  & 72.0  & 71.7  & 68.7  & 93.3  & 67.0  & 67.5  & 64.4  & 69.3 {\tiny ($\mathrel{+}$1.1)} \\
    \midrule
    \multirow{4}[2]{*}{AGNews} & Standard & 92.0  & 57.2  & 54.5  & 52.1  & 92.8  & 61.4  & 59.2  & 55.3  & 94.7  & 71.8  & 70.1  & 50.2  & 59.1  \\
          & ADV-G & 91.1  & 87.0  & 83.9  & 83.1  & 92.9  & 89.0  & 88.2  & 87.3  & 94.5  & 86.7  & 85.9  & 83.5  & 86.1  \\
          & ADV-F1 & 91.2  & 84.7  & 82.3  & 82.3  & 92.9  & 87.3  & 87.1  & 84.9  & 94.0  & 85.5  & 84.2  & 81.0  & 84.4 {\tiny($\mathbin{-}$1.7)}  \\
          & ADV-F2 & 91.4  & 86.8  & 83.8  & 82.9  & 92.9  & 88.5  & 87.5  & 85.6  & 94.5  & 86.0  & 84.4  & 83.6  & 85.5 {\tiny($\mathrel{-}$0.6)}  \\
    \bottomrule
    \end{tabular}%
  \label{tab:main}%
\end{table*}%

\paragraph{$n$-FD Convex Hull}
In $n$-FD adversarial training, we replace the gradient ascending direction with frequency descend direction to generate virtual AEs, formulated as:
\begin{equation}
\label{eq:nfd w}
    \Delta w_j^{(i)} = - \alpha \left\lVert
    \Delta \Phi_n(s_j^{(i)};\boldsymbol{x}) \right\rVert,
\end{equation}
where $\Delta \Phi_n$ is defined in Eq.~(\ref{eq:freq_delta}).
This equation aims to increase the weight of $n$-FD substitutions. 
The full training algorithm is showed in Alg.~\ref{alg:nfd CV}.

We implemented two $n$-FD training methods based on 1-gram and 2-gram frequency, denoted as \textbf{ADV-F1} and \textbf{ADV-F2},  respectively\footnote{For memory and computational speed reasons, we didn't implement algorithm with larger $n$.}.
We follow the implementation of~\cite{DNE}.


When $n=1$, ADV-F1 updates $w_j^{(i)}$ by its corresponding word frequency, the $\Delta \Phi_1(s_j^{(i)}; \boldsymbol{x})$ in Eq.~\ref{eq:nfd w} is computed as follows:
\begin{equation}
\label{unigram}
    \Delta \Phi_1(s_j^{(i)}; \boldsymbol{x}) = \phi(s_j^{(i)}) .
\end{equation}
Notice that we omit the frequency of the original example $\Phi_n(\boldsymbol{x})$ as it remains constant.
When $n=2$, ADV-F2 use the frequency of two 2-grams that contains $s_j^{(i)}$ to update $w_j^{(i)}$.
$\Delta \Phi_2(s_j^{(i)}; \boldsymbol{x})$ is computed as follows:

\begin{equation}
    \label{bigram}
    \Delta \Phi_2(s_j^{(i)}; \boldsymbol{x}) = \phi([x_{i-1}, s_j^{(i)}]) + \phi([s_j^{(i)}, x_{i+1}]).
\end{equation}

Notice that the frequency information is updated during training, so the update direction of $w$ is dynamic.

\subsection{Experimental Settings}
\label{sec:robust_exp}
\paragraph{Dataset and Models}

Dataset includes Internet Movie Database (IMDB)~\cite{IMDB} for sentiment classification task and the AG-News corpus (AGNews)~\cite{AGnews} for topic classification task. 
Experiments are adopted on three different DNN architectures: convolutional neural network (CNN)~\cite{wordcnn}, long short-term memory (LSTM)~\cite{lstm} and pre-trained BERT~\cite{bert}. 

\paragraph{Evaluation Metrics}
We use the following metrics to evaluate defensive performance:
1) \emph{Clean accuracy} (CLN) represents the model's classification accuracy on the clean test set. 
2) \emph{Robust accuracy} is the model's classification accuracy on the examples generated by a specific attack.
A good defender should have higher clean accuracy and higher robust accuracy.



\paragraph{Attacks}
We utilize three powerful word-level attacks to examine the robustness empirically: PWWS~\cite{PWWS}, TextFooler (TF)~\cite{textfooler} and LocalSearch (LS)~\cite{localsearch}. 
PWWS and TF employ greedy search with word-saliency exploring strategy. 
They first compute the word saliency of each original word and then do greedy substitution.
On the other hand, LS is an iterative attacker who selects the worst-case transformation at each step. Thus, LS achieve a higher successful attack rate but requiring more query numbers. 
For fair comparison, we use the same substitution set for all defenders, following the setting of \cite{DNE}.

\begin{figure}[t]
     \centering
     \centering
     \includegraphics[width=0.49\textwidth]{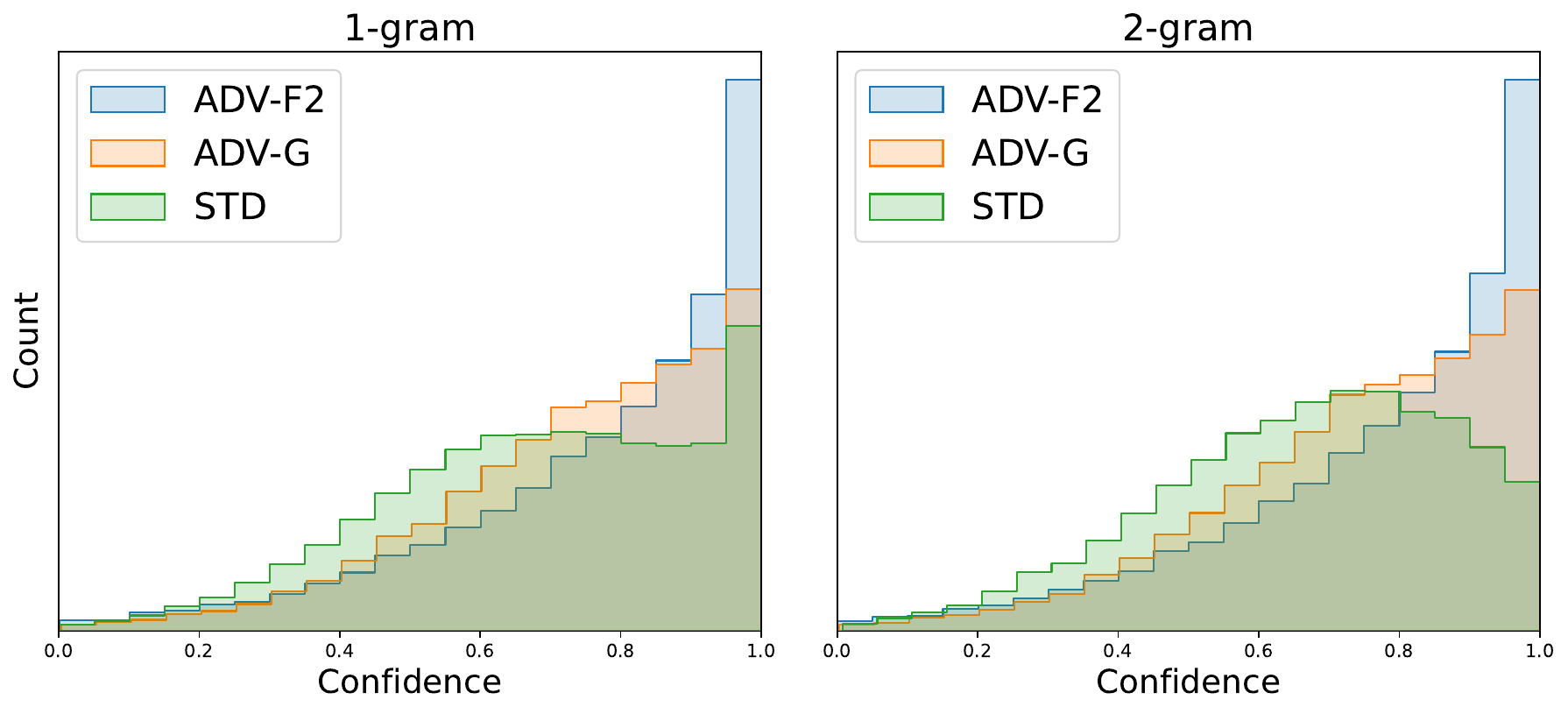}
     \caption{Confidence distribution of different models on $n$-FD examples.
     After training on AEs, models also achieve better performance on $n$-FD examples.
     }
     \label{fig:conf dist rob}
\end{figure}

\subsection{Results and Analysis}
\textbf{Training models on $n$-FD examples improves robustness.}
Table~\ref{tab:main} reports the clean accuracy (CLN) and robust accuracy against three attacks (PWWS, TF, LS) across two dataset. 
We observe that both ADV-G and ADV-F effectively enhance the model's robustness.
ADV-F achieves competitive defensive performance with ADV-G, with only a minor difference of less than 2\% 
This small performance gap suggests that adversarial examples generated by both gradient methods and $n$-gram frequency have similar impacts on model robustness.
Another key finding is that ADV-F2 consistently outperforms ADV-F1, indicating that 2-FD examples are more effective in increasing robustness than 1-FD examples.
This observation aligns with the earlier findings discussed in Section~\ref{sec:attack result}.
Further analysis on how the choice of $n$-value influences robustness enhancement in Section~\ref{sec:n_discussion}.

\begin{figure}[t]
    \centering
    \includegraphics[width=0.49\textwidth]{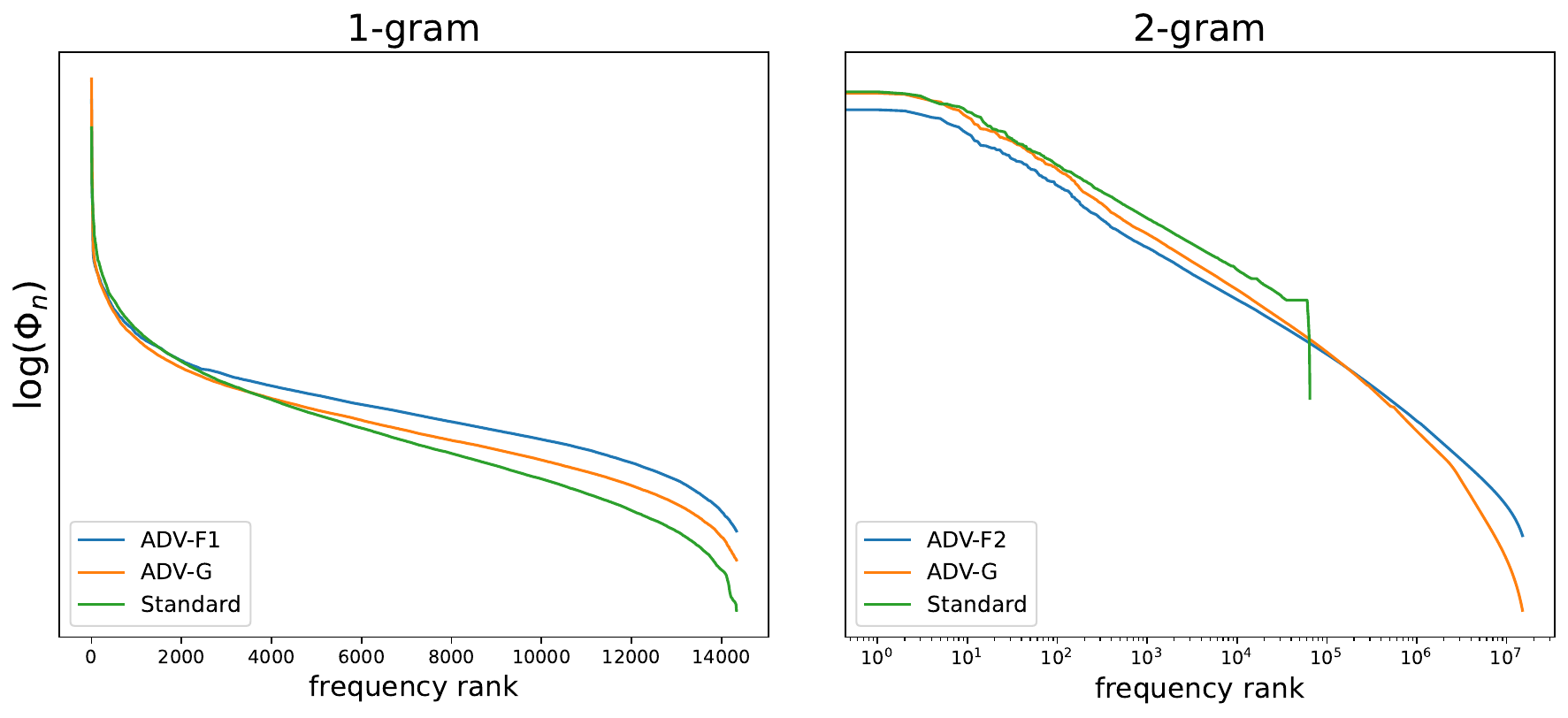}
    \caption{The $n$-gram frequency distribution of all training examples for ADV-G, ADV-F and standard training on IMDB. We normalize the frequencies with the training data size for comparison.Both ADV-G and ADV-F result in a more balanced $n$-gram frequency distribution, i.e., lower at the head but higher in the tail.}
    \label{fig:train_freq}
\end{figure}

\textbf{Gradient-based adversarial training generates $n$-FD examples.}
Figure \ref{fig:train_freq} shows the sorted frequency distribution of all training examples, including those from adversarial (ADV-G, ADV-F) and standard training methods. 
Since standard training utilizes only clean examples, its distribution is the same as that of the original training set. 
One can observe that ADV-G, like ADV-F, increases the frequency of the less common $n$-grams.

\textbf{Adversarial training improve model's performance on $n$-FD examples. }
Figure \ref{fig:conf dist rob} displays the distribution of confidence scores for the correct class across various models when handling $n$-FD examples. 
These models are trained using standard or adversarial methods on the IMDB dataset.
Notably, after adversarial training, there is an increase in the confidence scores . 
Furthermore, both frequency-based and gradient-based training strategies effectively improve the model's performance on $n$-FD examples, aligning with our expectations. 

\subsection{Exploration of Proper $n$ for Robustness Improvement}
\label{sec:n_discussion}
\begin{figure}[t]
    \centering
    \includegraphics[width=0.48\textwidth]{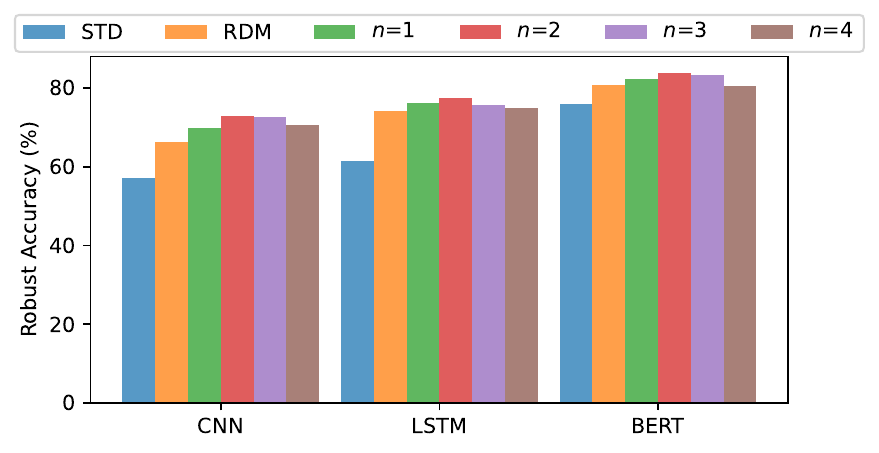}
    \caption{The robustness performance of $n$-FD example augmentation with different $n$-values. ``STD'' is short for standard training, and ``RDM'' is short for random augmentation.}
    \label{fig:n_robust}
\end{figure}

We conduct another defensive experiment to explore the robustness improvement considering different value of $n$. 
We generate $n$-FD examples of each training example for different $n$ values, and then augment them to the training set. 
The results in Figure~\ref{fig:n_robust} show the different robustness performance when $n$ increases from 1 to 4.
The robust accuracy is measured by PWWS attack~\cite{PWWS} on 1000 examples from AG's News dataset, across three model architectures.
We can observe that  the optimal defense performance occurs when $n$ is 2, with a decline in performance as $n$ increases.
This trend suggests that 2-FD information more effectively finds examples that enhance model robustness.
However, as $n$ become larger, the $n$-FD  loses its informational value. Because most substitution will be considered as $n$-FD,  reducing the process to random augmentation.

\section{Related Works}

\subsection{Textual Adversarial Attack}

Despite the great success in the NLP field, deep neural networks are shown to be vulnerable against AEs in many NLP tasks~\cite{eg:read,eg:class,eg:NMT,li2023vrptest, lu2023large, li2022unleashing, yang2024reducing, jiahao2023DConRec, wu-et-al:DcRec_cikm22, liu2023large}.
Textual adversarial attacks can be classified by granularity into character-level, word-level, sentence-level and mixture-level~\cite{attacksurvey}.
Character-level attacks focus on deleting, adding or swapping characters~\cite{deepwordbug,textbugger}, which usually results in grammatical or spelling errors~\cite{PruthiDL19}. 
Sentence-level attacks change a whole sentence to fool the victim model, e.g., paraphrasing or appending texts~\cite{AddSent}. And mixture-level attacks combine different level operations, e.g. phrases and words~\cite{lei2022phrase, Guo2021TowardsVT}.
In comparison, word-level attacks craft AEs by modifying words, where the candidates are formed by language databases~\cite{wordnet,hownet}, word embeddings~\cite{CF} or large-scale pre-trained language modeling.
These word-level attacks directly leverage gradient~\cite{hotflip}, search methods~\cite{GA, localsearch, MO} to find effective word substitutions.


\subsection{Textual Adversarial Defense}
The goal of adversarial defense is to make models have high performance on both clean and adversarial examples. 
Defense methods are generally categorized into empirical and certified, based on whether they provide provable robustness. 
Adversarial training and adversarial data augmentation are two popular approaches in empirical defense~\cite{freelb, PWWS, textfooler, dong1, dong2}.  
Adversarial training generates perturbation during training, while adversarial data augmentation obtains it after training, hence requiring a re-train phase.
However, such augmentation is insufficient due to the large perturbation space, so these methods cannot guarantee the robustness of the model.
Convex hull-based defense is another approach of adversarial training~\cite{DNE, ascc}, which optimizes the model's performance over the convex hull formed by the embedding of synonyms. 
On the other hand, certified defense provides a provable robustness.
Certified defense mainly consists of two types: Interval Bound Propagation (IBP)~\cite{ibp1,ibp-trans} and random smooth~\cite{safer}. 
IBP-based method computes the range of the model output by propagating the interval constraints of the inputs layer by layer, which requires knowing the structure of each layer. 
Random smooth methods achieve certified robustness by the statistical property of noised inputs.

To the best of our knowledge, only one previous work has explored frequency changes in  AEs~\cite{freqdetect}.
However, our research diverges significantly in several key areas, clearly establishing its distinct contribution to the field: 
1) Scope of analysis: The previous work concentrates on the single words. In contrast, our work embraces a broader scope, examining general $n$-gram frequency. Notably, our statistical analysis show that 2-grams provide more insightful results than single-word analysis.
2) Purpose and application: The previous work primarily utilized word frequency as features to detect attacks. Conversely, we employ frequency analysis as a tool to deepen our understanding of word-level attacks. Furthermore, we verified that the use of
$n$-FD examples specifically to improve the robustness of models.


\section{Conclusion}
This paper provides a novel understanding of word-level textual attacks through the lens of $n$-gram frequency, and provides a new direction to improve model robustness.
Our analysis of adversarial examples reveals a the attackers' general tendency towards $n$-FD examples, with $n = 2$ shows the .
We also find that typically trained models are more vulnerable to n-FD examples, indicating potential risks for NLP models.
Motivated by these findings, we introduce an n-FD adversarial training method that significantly improves model robustness, comparable to gradient-based approach.
Notably, using 2-gram frequencies proves more efficient in fortifying models compared to 1-gram frequencies.
We believe this work will deepen the understanding of adversarial attack and defense in NLP.
However, there are limitations in our study. Primarily, we do not fully understand why there are some AEs that belongs to 2-FA.
Furthermore, our study does not incorporate multiple n-gram information.

\section{Ethic Statement}

While our interpretations and experimental findings have the potential to design a more powerful attack, this recognition raises critical ethical concerns, as advancements in attack strategies could be misused, leading to more effective ways of deceiving NLP systems. However, it is crucial to emphasize that our primary objective is to contribute positively to the field by enhancing the understanding and defense mechanisms against such attacks. 
Moreover, we proposed a method specifically designed to improve the robustness of models against several attacks.

\section*{Acknowledgment}

This work was supported in part by Guangdong Major Project of Basic and Applied Basic Research (Grant No. 2023B0303000010), and in part by the National Natural Science Foundation of China under Grant 62272210.

\bibliographystyle{IEEEtran}
\bibliography{anthology}

\end{document}